\documentclass{llncs}  

\usepackage{graphicx} 
\usepackage[T1]{fontenc}
\usepackage{amsmath}
\usepackage[english]{babel}
\usepackage{array,tabularx}
\usepackage{booktabs} 
\usepackage{makeidx}  % allows for index generation
\usepackage{xparse}
\usepackage{moredefs}
\usepackage{lips}
\usepackage{epstopdf} 
\usepackage{color}
\usepackage{csquotes}
\usepackage{xcolor}
\usepackage{times}
\usepackage[hidelinks]{hyperref}
\usepackage[nolist,nohyperlinks]{acronym}
\usepackage{comment}
\usepackage{paralist}
\usepackage{cleveref}
\usepackage{wrapfig}
\usepackage{multirow}
\usepackage[flushleft]{threeparttable}
\usepackage{microtype}
\usepackage{listings}
\usepackage[super]{nth}

\usepackage{float}
\floatstyle{plaintop}
\restylefloat{table}

\usepackage{algorithm}
\usepackage{algorithmic}
\usepackage{amssymb}

\usepackage[style=base,labelfont=bf,labelsep=period,tableposition=top,font=small]{caption}
\usepackage{subcaption}
\makeatletter
\renewcommand\@biblabel[1]{#1.}
\makeatother
\addto\captionsenglish{}

\usepackage{fancyhdr}
\fancypagestyle{WI_footer}{\fancyhf{}\fancyfoot[L]{\color{black}\scriptsize 17th International Conference on Wirtschaftsinformatik\\February 2022, Nürnberg, Germany}}

\usepackage{cite}

\crefformat{footnote}{#2\footnotemark[#1]#3}
\expandafter\def\expandafter\UrlBreaks\expandafter{\UrlBreaks%  save the current one
  \do\a\do\b\do\c\do\d\do\e\do\f\do\g\do\h\do\i\do\j%
  \do\k\do\l\do\m\do\n\do\o\do\p\do\q\do\r\do\s\do\t%
  \do\u\do\v\do\w\do\x\do\y\do\z\do\A\do\B\do\C\do\D%
  \do\E\do\F\do\G\do\H\do\I\do\J\do\K\do\L\do\M\do\N%
  \do\O\do\P\do\Q\do\R\do\S\do\T\do\U\do\V\do\W\do\X%
  \do\Y\do\Z}

\newcolumntype{L}[1]{>{\raggedright\arraybackslash}p{#1}}   % linksbündig mit Breitenangabe
\newcolumntype{C}[1]{>{\centering\arraybackslash}p{#1}}     % zentriert mit Breitenangabe
\newcolumntype{R}[1]{>{\raggedleft\arraybackslash}p{#1}}    % rechtsbündig mit Breitenangabe

\begin{document}
\frontmatter          % for the preliminaries

\mainmatter              % start of the contributions

\title{No Time Like the Present: Effects of Temporal Language Change on Automated Comment Moderation} 

%\author{Blinded\inst{1}}
%\authorrunning{Blinded et al.} % abbreviated author list (for running head)
%\institute{Blinded}
%Deadline 05.11.2021
\author{Lennart Justen\inst{1} \and
Kilian Müller\inst{2} \and
Marco Niemann\inst{2}}
%\author{Authors blinded for review}
%\institute{Institutes blinded for review}
\institute{University of Wisconsin–Madison, Department of Physics, Madison, WI, USA\\
\email{ljusten@wisc.edu} \and
European Research Center for Information Systems (ERCIS), Münster, Germany\\
\email{\{marco.niemann, kilian.mueller\}@ercis.uni-muenster.de}}

% -----------------------
% |  Begin of Document  |
% -----------------------
\maketitle
\setcounter{footnote}{0}

% ------------- 
% |  Abstract and Keywords  |
% -------------
\begin{abstract}
The spread of online hate has become a major problem for newspapers that host comment sections. As a result, there is growing interest in using machine learning (ML) and natural language processing (NLP) for (semi-) automated abusive language detection to avoid manual comment moderation costs or having to shut down comment sections all together. However, much of the past work on abusive language detection with ML uses random train-test splitting procedures that assume an unrealistically static language environment. In this paper, we show using a new German newspaper comments dataset that a time-stratified evaluation procedure provides a more realistic measure of a classifier's performance on future data. We also show that the performance of classifiers can degrade quickly as the training data grows more outdated and language and news coverage evolve. Further, we demonstrate that the performance of classifiers trained on data from before the COVID-19 pandemic drops sharply when evaluated on COVID-era comments. Our findings suggest that when standard ML techniques are applied naively to abusive language detection, a classifier will fail to meet the advertised evaluation benchmarks in the real-world environment.

\end{abstract}

\thispagestyle{WI_footer}

% ------------- 
% |  Content  |
% -------------

\section{Introduction}
\label{sec:introduction}

In recent years abusive language and hate speech have become pernicious problems for online communication platforms, including newspapers, that host comment sections \cite{green_no_2018,gardiner_dark_2016}. 
At their inception, comment sections were welcomed by newspapers as a way to cultivate critical discourse between journalists and their audience and generate traffic (cf., \cite{Reich2011,Papacharissi2004}). 
In 2008, upon opening comment sections for their articles, the major U.S. media platform National Public Radio\footnote{\url{www.npr.org}} (NPR) wrote \enquote{We are providing a forum for infinite conversations on NPR.org. Our hopes are high. We hope the conversations will be smart and generous of spirit. We hope the adventure is exciting, fun, helpful and informative} \cite{meyer_npr_2008}. 
In 2016, inundated with trolls and toxic comments, NPR like many other major newspapers shut down their comment sections \cite{montgomery_beyond_2016,green_no_2018,gardiner_dark_2016}. 
Newspapers that opt to keep comment sections open have faced increasing legal scrutiny and significant comment moderation costs \cite{green_no_2018,gardiner_dark_2016,niemann_what_2020}.

In response to the increase in toxic comments and the associated costs, there has been growing interest in the field of abusive language detection (ALD). ALD is an application of machine learning (ML) and natural language processing (NLP) that can be used to develop (semi-) automated comment moderation tools \cite{salminen_developing_2020,brunk_can_2019,niemann_abusive_2020}. The realization that it would be necessary to (semi-)automate at least parts of the moderation workflow came early on \cite{Paulussen2011}, and work on the corresponding systems began more than a decade ago \cite{Yin2009}.

Against the backdrop of the refugee crisis in 2015, the field of abusive comment detection experienced a surge following Nobata et al.'s influential paper \cite{nobata_abusive_2016}.
Since then, various scholars around the world have primarily focused on creating machine learning models that could improve the detection quality, which is reflected in a growing number of (meta-) studies \cite{fortuna_survey_2018,niemann_abusive_2020,Pamungkas2021,Yin2021}.
Linked but typically subordinate to the machine learning work individual contributions cover related topics such as labels \cite{niemann_abusive_2020}, datasets \cite{vidgen_directions_2021,Poletto2020}, or the integration into moderation platforms \cite{Loosen2017,Niemann2021}.
However, not only academics have been working on a resolution to this pressing issue.
The New York Times\footnote{\url{www.nytimes.com}} (NYT) for example, one of the biggest U.S. newspapers, partnered with Alphabet\footnote{\url{https://abc.xyz/}} to build a tool called Perspective\footnote{\url{www.perspectiveapi.com}} to automatically flag toxic comments \cite{lecher_alphabet_2017}. 
At its roll-out in 2017, Perspective was reportedly already automatically approving about 20\% of the total comments received by the NYT \cite{lecher_alphabet_2017}.

However, despite all these efforts, the core problem---the effective, efficient, and ideally error-free automated detection of abusive comments--- has still not been solved, as many classifiers are showing fairly good but often exaggerated performances \cite{Yin2021}.
While this may be largely attributable to the complexity of the classification task and mistakes or imprecisions in the experimental setups, another---so far largely unconsidered---reason might lie within the data and the notion of language itself.  

Typically, the dataset is randomly split into training and testing subsets to ensure that the results are a reflection of the classifier's performance on future data \cite{sogaard_we_2020}. However, classifiers evaluated in this random train-test fashion tend to yield overoptimistic results so that when the model is deployed, it performs worse than expected. An important reason for this drop in performance is that the random train-test split falsely assumes a static language environment \cite{vidgen_challenges_2019,sogaard_we_2020,lazaridou_pitfalls_2021}. When split in this fashion, the training and testing data share the same time period. This condition does not hold when the model is deployed. The goal of a classifier deployed is to predict \textit{future} data, but a classifier trained under the false assumption that language is static will rely on a training corpus that is increasingly less representative of future data

Language and the subjects of our online commentary are constantly in flux. Some words that were once considered to have negative meanings like "wicked" and "sick" can now have positive meanings, and vice versa \cite{zeitlin_11_2019}---processes of semantic change known as amelioration and pejoration respectively \cite{frermann_bayesian_2016,lukes_sentiment_2018,cook_automatically_2010}. Language in online spaces like comment forums often changes especially rapidly as different forms of "netspeak" (i.e. internet slang) emerge and fade in social networks \cite{goel_social_2016,eisenstein_diffusion_2014}. The people, places, topics, and trends in newspaper articles are also constantly changing. For example, during the COVID-19 pandemic, words like \textit{coronavirus}, \textit{social distancing}, and \textit{quarantine} abruptly become widespread in news coverage. The wider phenomenon of dynamically changing data streams is known as \textit{concept drift} in computer science literature. In light of temporal changes in language, it is critical to determine how abusive language detection systems degrade over time.

This paper examines the temporal effects on the performance of abusive language detection classifiers trained on a German news comment dataset from Nov. 2018 to Jun. 2020. Our goals are to determine if random splits tend to overestimate model performance compared to time stratified evaluation, and measure \textit{temporal degradation}; whether and by how much the performance of a model decreases as the time between the training and testing data increases) \cite{lazaridou_pitfalls_2021}. 

The contribution of this paper is to caution practitioners of abusive language detection about the problem of concept drifting data and to provide evidence of performance degradation when standard ML techniques are applied naively. Newspapers that implement semi-automated ALD systems must be aware that maintaining the model's advertised performance benchmarks will require re-training on new data or implementing other, more complex adaption strategies.

The rest of the paper is structured as follows. Sec. \ref{sec:related_work} describes the phenomenon of concept drift and explores past findings in NLP applications. A detailed review of concept drift adaptation strategies is outside of the scope of this paper, but we reference several recognized review papers. Sec. \ref{sec:dataset} describes our dataset of German-language newspaper comments. Sec. \ref{sec:experimental_setup} describes two experiments we conducted to examine the temporal dynamics of ALD classifiers as well as our text-preprocessing techniques and automated machine learning (Auto-ML) approach to model development. In Sec. \ref{sec:experimental_results}, we report the results of our experiments and provide further evidence of temporal changes in language in our German newspaper comment dataset. Finally, in Sec. \ref{sec:discussion} we outline some of the implications of our findings for practitioners of ALD and provide directions for future investigation. 

\section{Related Work}
\label{sec:related_work}
In many machine learning applications, models need to adapt to dynamic environments where the incoming data or the target outputs change unpredictably. In this section, we will describe the problem of changing data as the \textit{concept drift} problem and summarize some of the most relevant NLP papers that deal with concept drift and temporal degradation. 

\subsection{Concept Drift}
Comments posted to a newspaper website arrive as a data stream (although data streams are sometimes differentiated by the high rate of speed and real-time or "one-pass" analysis)\cite{bifet_machine_2018}. Temporal changes in a data stream, i.e., changes in the mapping between the input data and the target output variable across time, are a well-studied phenomenon called concept drift \cite{dries_adaptive_2009,kifer_detecting_2004,gama_survey_2014}. Concept drift between time $t_0$ and $t_1$ can be written as

\begin{equation}
    \exists X: p_{t_{0}}(X, y) \neq p_{t_{1}}(X, y)
\end{equation}

where $p_{t_0}$ is the joint distribution at time $t_0$ between the set of input variables $X$ and the target variable $y$ \cite{gama_survey_2014}.

There are two different types of concept drift: \textit{real} concept drift and \textit{virtual} concept drift. Real concept drift refers changes in $p(y|X)$ that occur either with or without changes in $p(X)$ \cite{gama_survey_2014}. In abusive comment moderation, real concept drift is a change in the types of content moderators consider abusive. These changes may occur independently of changes in the incoming data like new hate speech regulations being put into effect or a managerial decision to raise the threshold for restricting comments. 

Virtual concept drift describes changes in distribution of the incoming data $p(X)$ without affecting $p(y|X)$ \cite{gama_survey_2014}. In abusive comment moderation, virtual concept drift is the more likely type of drift as the language of abuse and the targets of abuse (e.g., people, places, organizations) change. Virtual drift also encompasses changes in the class distribution (i.e., changes in the proportion of abusive comments and clean comments).

There has been growing interest in the domains of \textit{concept drift detection} and \textit{adaptive learning} to develop concept drift adaption strategies. Adaptive learning is a concept drift adaption strategy in which the model is updated online during operation \cite{gama_survey_2014}. A wide variety of adaptive learning algorithms have been proposed ranging from simple sliding-window techniques in which older data is gradually dropped from the training corpus to complex learning algorithms that combine drift detection methods with sophisticated data forgetting mechanisms. For the most comprehensive review of concept drift adaptation strategies, we refer readers to Gama et al. 2014 \cite{gama_survey_2014} and Tsymbal 2004 \cite{Tsymbal_2004}. 

\subsection{Drift and Degradation in NLP Tasks}

The problem of concept drift and temporal degradation has been studied in a variety of NLP tasks including document classification \cite{lazaridou_pitfalls_2021,huang_examining_2018,rocha_exploiting_2008,silic_exploring_2012}, sentiment analysis \cite{lukes_sentiment_2018,muller_addressing_2020,bifet_sentiment_2010,a_bechini_addressing_2021}, named entity recognition \cite{rijhwani_temporally-informed_2020}, fake review detection \cite{mohawesh_analysis_2021}, spam filtering \cite{delany_case-based_2005}, and abusive language detection \cite{nobata_abusive_2016,florio_time_2020}. In the longest time interval studied, \cite{huang_examining_2018} classified sentences from American political platforms between 1948 and 2016 as either Democrat or Republican. By training and testing on data from different time intervals, they showed that F1-scores degraded by 40 points in some cases. Others like \cite{lazaridou_pitfalls_2021,muller_addressing_2020,rijhwani_temporally-informed_2020,florio_time_2020} have shown that classifiers can degrade over much shorter periods on the order of months. 

In most cases like \cite{lazaridou_pitfalls_2021,silic_exploring_2012,a_bechini_addressing_2021,rijhwani_temporally-informed_2020,florio_time_2020} this degradation accumulates steadily over time due to \textit{gradual} concept drift in the data \cite{gama_survey_2014}. However, in some cases, an abrupt change in the data (i.e. \textit{abrupt} concept drift \cite{gama_survey_2014}) can cause a sudden drop in performance. For example, \cite{muller_addressing_2020} examined concept drift in vaccine-related sentiment analysis and found that the performance of outdated classifiers suddenly dropped by about 20\% in the early months of 2020 when the COVID-19 pandemic began to receive global attention. 

To the best of our knowledge, Florio et al. \cite{florio_time_2020} is the only other work to measure temporal degradation in the context of abusive language detection. They trained two models---a Support Vector Machine (SVM) and Google's Bidirectional Encoder Representation Transformer (BERT) \cite{devlin_bert_2019} adapted for Italian---on an Italian language Twitter dataset of 4,000 samples from 2015 to 2017. Both models were then evaluated on monthly evaluation datasets of 2,000 samples from Sept. 2018 to Feb. 2019. Within the six months of evaluation data, the BERT and SVM models lost 0.227 and 0.284 F1 points, respectively---a drop that would severely cripple the usability of these models in a real-world setting.  

Although past findings of concept drift indicate its relevance to ALD systems, our study contributes several novel insights. All languages undergo temporal changes. However, it is not clear how these processes manifest in different languages. A review by Niemann et al. \cite{niemann_abusive_2020} found that German datasets for abusive language detection are both relatively rare (compared to English, Italian, and Indonesian) and tend to yield worse F1 scores. Thus our dataset provides a unique perspective on the problem of concept drift in a German ALD setting. Our data also overlaps the emergence of the COVID-19 pandemic, which was accompanied by intense news coverage and changes in our vocabulary. The period of our data allows for an examination of the concept drift associated with the pandemic. Overall, this work aims to integrate insights from concept drift literature and the field of ALD, an intersection that is so far poorly understood.

\section{Dataset}
\label{sec:dataset}

The basis for the conducted experiments (cf., Sec. \ref{sec:experimental_setup}) is provided by an extensive German news comment dataset. 
The dataset was provided by one of the largest German newspapers and contains comments submitted to the newspaper's website by the readers. 
%The aforementioned newspaper utilizes pre-moderation. During pre-moderation every submitted comment is evaluated by the community managers of the newspaper, whether it is acceptable to be published on their website. 
To provide a safe discussion space and to prevent legal trouble, each incoming comment is checked by a team of professional community managers before being published (pre-moderation process, cf., \cite{Reich2011,Grimmelmann2015}).

\begin{table}[!h]
\centering
    \begin{tabular}{lr}
        \toprule
        \textbf{Characteristic} & \textbf{Value}\\\midrule
        Number of comments &  256,173\\
        Number of accepted comments & 239,323\\
        Number of rejected comments & 16,850\\
        \bottomrule
    \end{tabular}
    \caption{Dataset Characteristics}
    \label{tab:data_characteristics}
\end{table}

If a comment is deemed non-publishable (e.g., by containing racist or sexist content), it will not appear on the website. 
%However, within the dataset, these problematic comments are contained.
However, differing from scraped datasets, all comments are included, even those too critical to be published.
As depicted in Tab.~\ref{tab:data_characteristics}, the dataset consist of more than 250,000 comments, around 17,000 ($\approx$6.5\%) were rejected by moderators, the remaining 240,000 ($\approx$93.5\%) were considered non-problematic. Each instance within the dataset is represented as shown in Tab.~\ref{tab:data_values}. Each entry contains a unique identifier (ID) as well as the date and time the readers of the newspaper posted the comment. All data within the dataset originates from user comments within a timeframe starting at the 01.11.2018 and ending on the 29.06.2020. Further entries are the textual content of each comment as well as the resulting moderation decision (comment was accepted by moderators results in "0", comment was rejected by moderators results in "1"). Lastly, each comment's length (number of characters used) is listed.

\begin{table}[!h]
    \centering
    \begin{tabular}{llcc}
        \toprule
        \textbf{Column} & \textbf{Description} & \textbf{Datatype} & \textbf{Ranges}\\
        \midrule 
        ID & Unique Identifier & \texttt{int} & - \\
        Date & Date and time the comment was posted & \texttt{datetime} & 01.11.2018 - \\
        &&&29.06.2020\\
        Text & Text of the comment & \texttt{text} & -\\
        Rejected & Decision if the comment is rejected by  & \texttt{bool} & [0,1]\\ 
        & the moderator &&\\
        Comment\_length & Number of characters within a comment & \texttt{int} & [0,26516]\\
        \bottomrule
    \end{tabular}
    \caption{Dataset Values}
    \label{tab:data_values}
\end{table}

% \begin{table}
% \centering
%     \caption{Dataset Characteristics and Structure}
%     \label{tab:data_characteristics}
%     \footnotesize
    
%     \begin{subfigure}[c]{1\textwidth}
%     \centering
%     \begin{tabular}{lr} 
%         \toprule
%         Characteristic & Value\\\midrule
%         Number of comments &  256173\\
%         Number of accepted comments & 239323\\
%         Number of rejected comments & 16850\\
%         \bottomrule
%     \end{tabular}
%     \end{subfigure}
    
%     \vspace{0.15cm}
    
%     \begin{subfigure}[c]{1\textwidth}
%     \centering
%     \begin{tabular}{llcc} 
%         \toprule
%         Column & Description & Datatype & Ranges\\
%         \midrule 
%         ID & Unique Identifier & \texttt{int} & - \\
%         Date & Date and time the comment was posted & \texttt{datetime} & 01.11.2018 - \\
%         &&& 29.06.2020\\
%         Text & Text of the comment & \texttt{text} & -\\
%         Rejected & Decision if the comment is rejected by & \texttt{bool} & [0,1]\\
%         & the moderator &&\\ 
%         Comment\_length & Number of characters within a comment & \texttt{int} & [0,26516]\\
%         \bottomrule
%     \end{tabular}
%     \end{subfigure}

% \end{table}

\section{Experimental setup}
\label{sec:experimental_setup}

This section describes the experimental setup for two tests: the time-stratified vs. random split test and the temporal degradation test. In the time-stratified vs. control experiment, we use a time-stratified evaluation procedure to examine how random train-test splits in concept drifting data can result in overoptimistic measures of performance. In the temporal degradation test, we sequentially chunk our dataset and measure how classifier performance depends on the time interval between training and testing data. We also describe our preprocessing procedure and our Auto-ML approach to model selection.

\subsection{Preprocessing}
\label{sec:preprocessing}
Before being transformed into numerical features, all text was preprocessed, including stopword removal, lemmatization, and lower-casing %\footnote{To view the complete preprocessing procedure see the \hyperlink{https://github.com/lennijusten/MODERAT/blob/main/TextPreprocessingTransformer.py}{TextPreprocessingTransformer} class on GitHub: \url{https://github.com/lennijusten/MODERAT}}
. The preprocessed text was then numerically represented with Term Frequency — Inverse Document Frequency (TF-IDF) vectors. In both the time-stratified vs. random split test and the temporal degradation test, we selected the top 3,000 most frequent unigrams and bigrams as TF-IDF features. We removed words that appeared less than five times in the dataset from the possible feature space. We used random undersampling across the entire moderator labeled dataset to achieve a balanced class distribution.

\subsection{Auto-ML Setup}
Finding the optimal ML configuration for a problem usually involves a repetitive and time-consuming process of testing different models, hyperparameters, preprocessing techniques, and feature engineering strategies. The goal of Auto-ML is to automate much of this workflow and reduce the developer's bias towards prioritizing specific models or configurations over others \cite{yao_taking_2019,jorgensen_multi-class_2020}.

In this paper we used the popular \textit{Auto-sklearn} Auto-ML library to train our classifiers \cite{Feurer2015}. Auto-sklearn is built on the well known \textit{scikit-learn} python ML library \footnote{See \url{https://automl.github.io/auto-sklearn/master/} for Auto-sklearn documentation} and uses Bayesian optimization methods across 15 classifiers, 14 feature preprocessing methods, 4 data preprocessing methods, and over 100 hyperparameters (model dependent) to construct an ensemble of the best performing models  \cite{Feurer2015}. In the time-stratified vs. random split test, we capped each Auto-sklearn instance at one day of run time with 20GB of available memory. In the temporal degradation test, we capped each Auto-sklearn instance at 12 hours of run time with 25GB of available memory. All 15 classifiers were included in the parameter search space for both experiments.

\subsection{Time-stratified vs. random split test}
% TODO: add schematic 

The time-stratified vs. random split experiment adapted from \cite{lazaridou_pitfalls_2021} compares the popular random train-test data split with a time-stratified train-test split. A time-stratified split means the model is trained on one time period of data and then evaluated on a subsequent time period with no overlap in the time period of the training and evaluation datasets. On the other hand, in a random split, the evaluation dataset is selected at random from the entire time period of data meaning there is total overlap in the time period of the training and evaluation datasets. The goal is to determine whether the random split tends to overestimate performance on future unseen data. 

In order to compare the two splitting strategies, we create an \textit{evaluation dataset} and two training datasets---the \textit{time-stratified dataset} and the \textit{control dataset}. The evaluation dataset contains all comments from the last eight months (i.e., Nov. 2019 through June 2020). The time-stratified dataset contains all comments from the first month until the evaluation period (i.e., Nov. 2018 to Nov. 2019). The control dataset contains all comments from the entire corpus period (i.e., Nov. 2018 to June 2020). In other words, the control dataset and evaluation dataset have overlapping time periods, whereas the time-stratified dataset and evaluation dataset do not. The model trained on the time-stratified dataset is evaluated on its ability to predict future comments posted after the time period of its training data. In contrast, the model trained on the control dataset is trained to predict comments posted during the time period of its training data.

\begin{figure}[htp]
    \centering
    \includegraphics[width=0.95\textwidth]{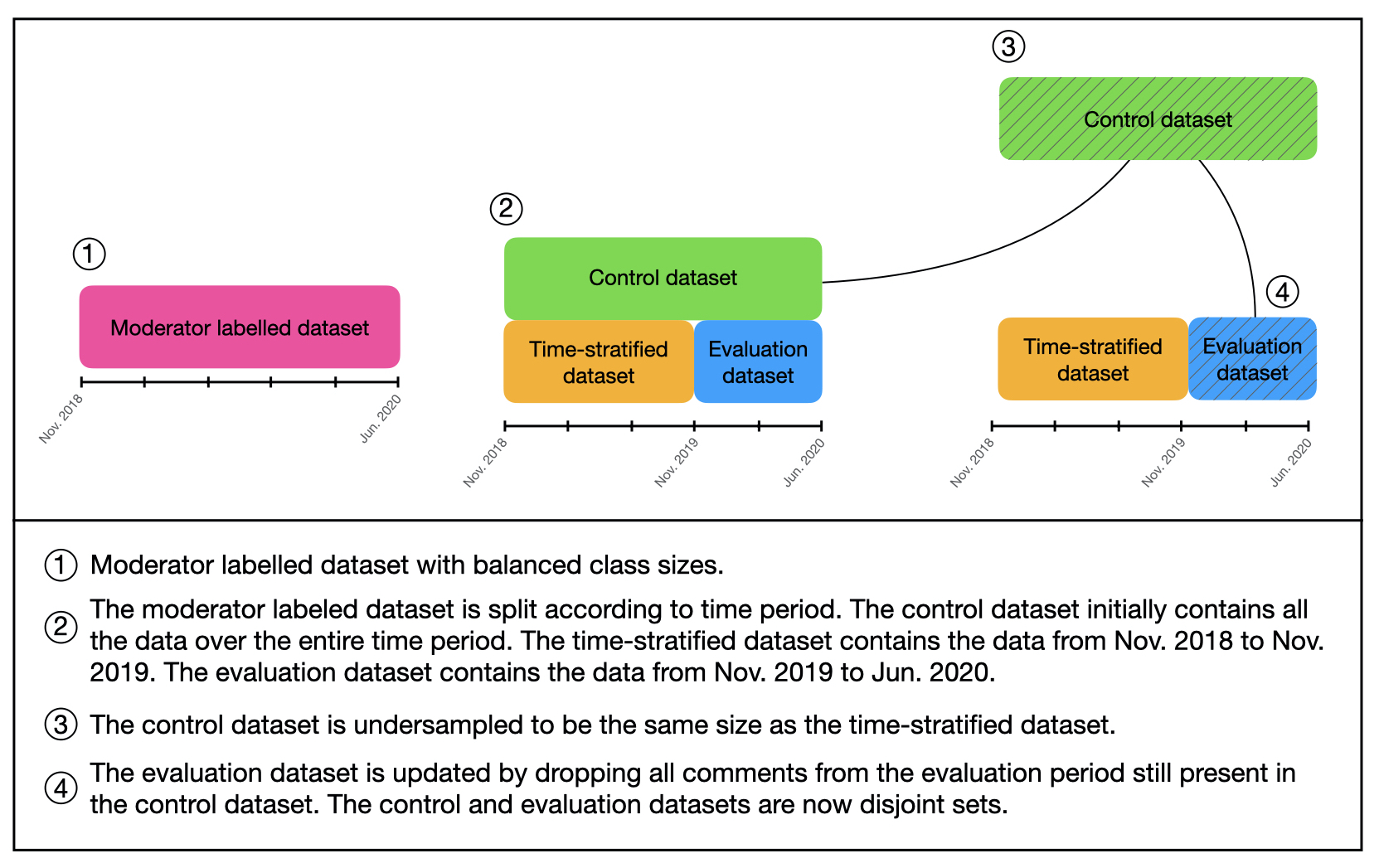}
    \caption{Procedure for creating the control, time-stratified, and evaluation datasets}
    \label{fig:flow}
\end{figure}

\begin{figure}[htp]
    \centering
    \includegraphics[width=0.95\textwidth]{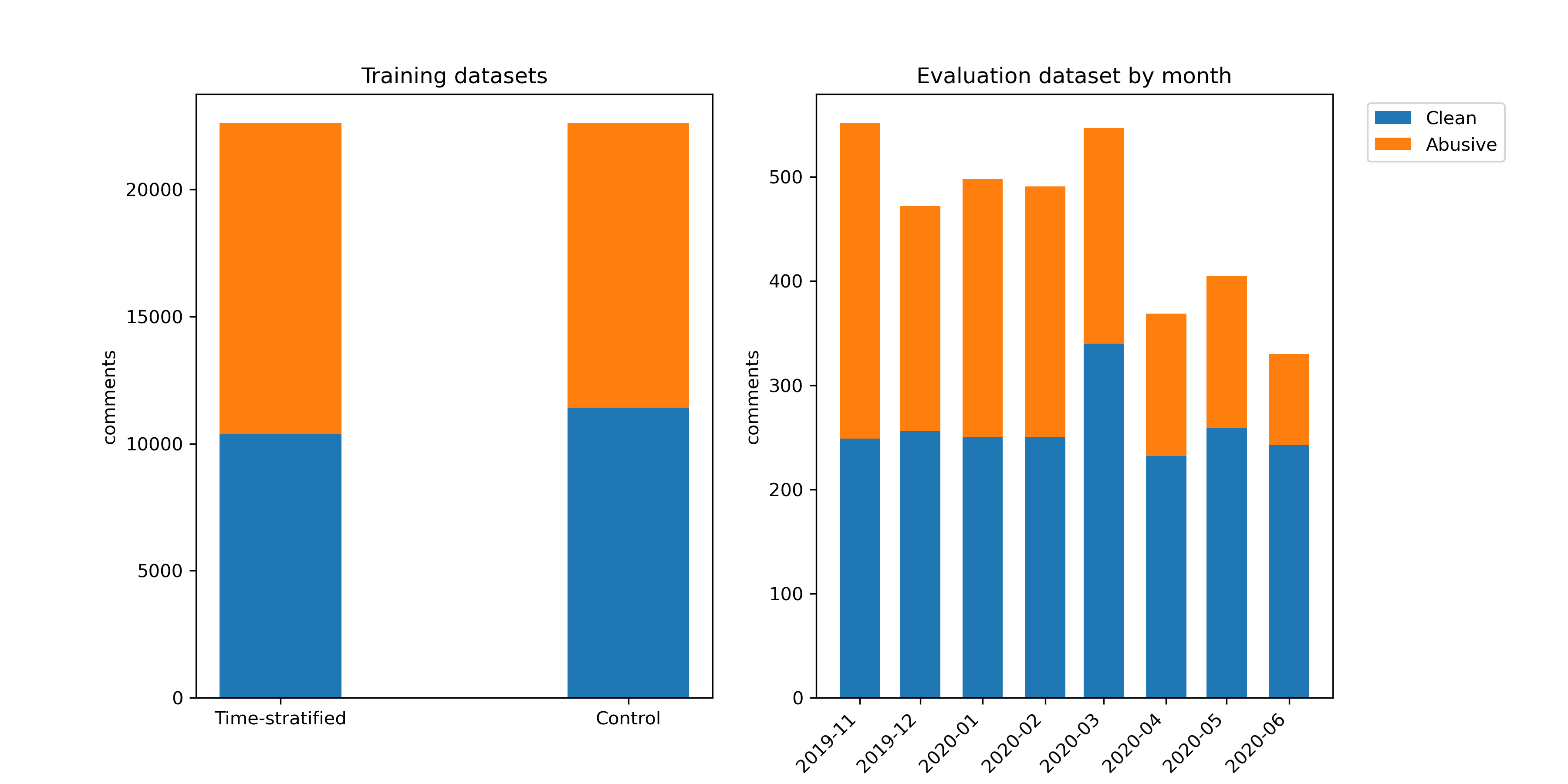}
    \caption{Overview of the control, time-stratified, and evaluation datasets}
    \label{fig:bar}
\end{figure}

The experiment has two additional requirements. First, the control and time-stratified training datasets must be of equal size. Second, no comments from the evaluation dataset are in either training dataset (i.e., the control-evaluation and time-stratified-evaluation pairs are disjoint sets). These requirements ensure that the two training datasets differ only in the time period of training data and that they are tested on the same evaluation dataset. We achieve this by randomly undersampling the control dataset to be the same size as the time-stratified dataset and then updating the evaluation dataset to exclude all comments present in the control dataset. The flowchart in Figure \ref{fig:flow} shows the four-stage process for creating the time-stratified, control, and evaluation datasets for the time-stratified vs. random split test. Figure \ref{fig:bar} shows the composition of clean and abusive comments in the training and evaluation datasets. The evaluation dataset is examined at a monthly resultion to assess how the performance of the control-trained classifier degrades as the time between the control period and the evaluation period increases.   

\subsection{Temporal degradation test}
In the temporal degradation test, we examine how the performance of a classifier changes over time. We measure temporal degradation by splitting the dataset into sequential chunks for training and evaluation and observing how classifier performance changes as the time interval between the training and evaluation data changes. The expectation is that as the time interval between training and evaluation data increases, the performance will degrade as language and news fluctuate.

In this experiment, we split the dataset into five consecutive four-month chunks. All chunks are then undersampled to match the number of comments in the smallest chunk (n=5007). Next, we fit an auto-sklearn classifier to each chunk and evaluate it on all other chunks. When training and testing on the same chunk, we use a random 20\% hold-out set for evaluation and the remaining 80\% for training.

To gain further insight into why performance varies, we also measure the corpus similarity between each chunk with the Spearman rank correlation coefficient \cite{kilgarriff_comparing_2001,mukaka_statistics_2012}. We calculate the Spearman coefficient with TFIDF-ranked word lists for the two corpora being compared\footnote{Python  \url{https://docs.scipy.org/doc/scipy/reference/generated/scipy.stats.spearmanr.html}}. A Spearman coefficient of 0 indicates no correlation, whereas a coefficient of 1 indicates perfect correlation. Given the temporal changes in language, we expect that the further apart two corpora are in time, the lower their correlation. 

\section{Experimental results}
\label{sec:experimental_results}

In this section, we describe the results of the time-stratified vs. random split test and the temporal degradation test outlined in Sec. \ref{Sec:experimental_setup}. We also provide further insights into the impact of the COVID-19 pandemic on classifier performance. In the time-stratified vs. random split test, we show that emerging new words during the evaluation period are dominated by pandemic-related vocabulary. Furthermore, in both tests, our time-stratified evaluation procedures show clear performance drops during the pandemic's early months.

\subsection{Time-stratified vs random split test}

\begin{figure}[htp]
    \centering
    \includegraphics[width=0.85\textwidth]{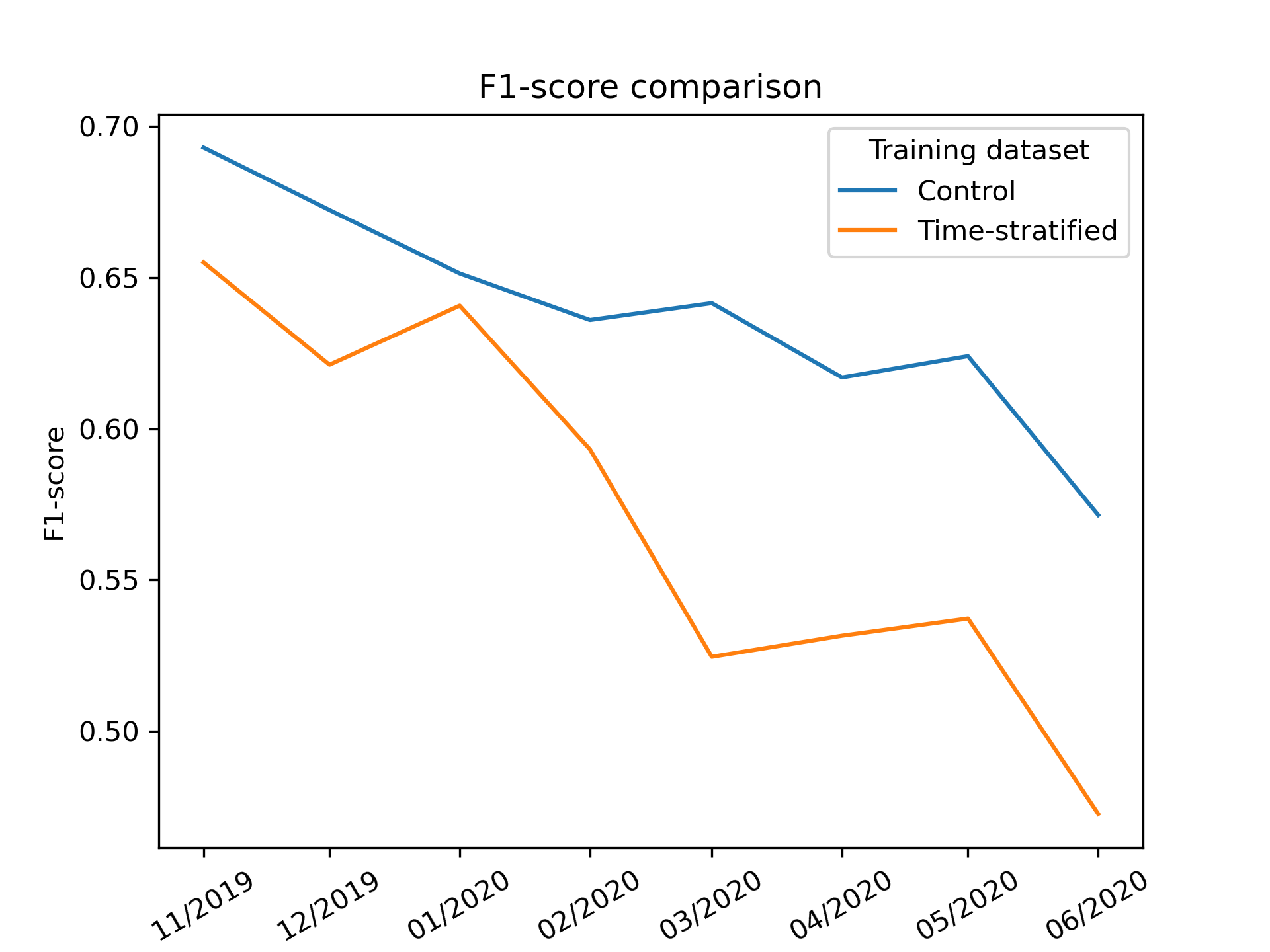}
    \caption{Comparison of two classifiers trained on the time-stratified versus control datasets}
    \label{fig:f1-compare}
\end{figure}

The results of our time-stratified vs. random split test confirm that a classifier evaluated with a random splitting technique (the control trained classifier) will yield overoptimistic results when compared to a time-stratified approach to evaluation. The classifier trained on the control dataset---the dataset which contains comments from the same time period as the evaluation dataset---had an overall F1-score of 0.632 compared with an F1-score of 0.590 for the classifier trained on the time-stratified dataset. An examination of the monthly performance across the evaluation dataset (Fig. \ref{fig:f1-compare}) shows that the control-trained classifier consistently performs better than the time-stratified-trained classifier. It also shows that the performance gap between the two classifiers grows as the time-stratified training data becomes increasingly outdated. Table \ref{tab:results} similarly shows the monthly performance of the two classifiers on the evaluation dataset with precision and recall data. In an abusive language detection environment, recall, or the number of abusive comments detected divided by the total number of abusive comments in the evaluation dataset, is critical. In a semi-automated abusive language detection system, the goal is to detect as many abusive comments as possible (high recall), even if that comes at the cost of classifying more clean comments as abusive (low precision) since moderators review comments marked as abusive.

% as the distance between the evaluation month and time-stratified dataset increases

\begin{table}[]
\centering
\caption{Comparing the performance of models trained on the control and time-stratified datasets}
\label{tab:results}
    \setlength{\tabcolsep}{12pt}
    \renewcommand{\arraystretch}{1.15}
    \footnotesize
    \begin{tabular}{ccccccc}
    \toprule
    \multirow{2}{*}{\textbf{Evaluation month}} &            \multicolumn{3}{c}{\textbf{Control}}                        &               \multicolumn{3}{c}{\textbf{Time-stratified}}             \\ \cmidrule(r){2-4} \cmidrule(l){5-7} 
                                               & \textbf{Prec.}         & \textbf{Rec.}        & \textbf{F1}            & \textbf{Prec.}          & \textbf{Rec.}        & \textbf{F1}           \\ \hline
    Nov. 2019                                  &         0.713          &       0.693          &          0.693         &        0.700            &       0.661          &         0.655         \\
    Dec. 2019                                  &         0.719          &       0.694          &          0.672         &        0.690            &       0.653          &         0.621         \\
    Jan. 2020                                  &         0.674          &       0.659          &          0.651         &        0.685            &       0.655          &         0.641         \\
    Feb. 2020                                  &         0.676          &       0.651          &          0.636         &        0.663            &       0.622          &         0.593         \\
    Mar. 2020                                  &         0.673          &       0.677          &          0.642         &        0.633            &       0.603          &         0.525         \\
    Apr. 2020                                  &         0.642          &       0.648          &          0.617         &        0.623            &       0.603          &         0.532         \\
    May 2020                                   &         0.658          &       0.665          &          0.624         &        0.630            &       0.613          &         0.537         \\
    Jun. 2020                                  &         0.635          &       0.666          &          0.572         &        0.604            &       0.605          &         0.473         \\ \bottomrule
    \end{tabular}
\end{table}

Interestingly, the performance of the control-trained classifier also degrades over time. One reason for this decline may be that the comments in the control dataset skew towards the earlier periods of the dataset---only 32.7\% of the control dataset comes from the evaluation period. The presence of old and outdated data in a training dataset can negatively impact classifier performance on new data, a phenomenon that runs counter the usual assumption that more data leads to better performance \cite{nobata_abusive_2016,gama_survey_2014}. Many concept drift adaption algorithms implement forgetting mechanisms like the sliding window in which old data outside of a particular time window is discarded \cite{gama_survey_2014,silic_exploring_2012}. 

We also note that during the onset of the COVID-19 pandemic in the early months of 2020, the time-stratified-trained classifier drops sharply (cf. Fig. \ref{fig:f1-compare}). This drop is likely due to the emergence of new vocabulary and abuse associated with the pandemic that was not present in the time-stratified dataset. Words that appear for the first time in the evaluation period (i.e., Nov. 2019 to June 2020) are "unseen" by a classifier trained on the time-stratified dataset but will likely have appeared in the control dataset. Table \ref{tab:words} shows a list of emerging words in the evaluation period. The list is dominated by frequently used words related to the COVID-19 pandemic. A classifier trained on a corpus of words without the pandemic-related vocabulary is almost guaranteed to run into trouble when faced with comments that, for example, decry lockdowns and vaccines, or blame certain groups for the emergence of the virus. As described in Sec. \ref{sec:related_work}, changes in the language and topics of abuse lead to concept drift, which standard ML models are poorly equipped to address. 

% Add briefly that usernames can be emerging new words
\begin{table}[]
\centering
\caption{Emerging new words in the evaluation period grouped by two-month intervals}
\label{tab:words}
\begin{threeparttable}
\footnotesize
\begin{tabular}{llllllll}
\toprule
\multicolumn{2}{c}{\textbf{11-12/2019}} & \multicolumn{2}{c}{\textbf{01-02/2020}} & \multicolumn{2}{c}{\textbf{03-04/2020}} & \multicolumn{2}{c}{\textbf{05-06/2020}} \\ \cmidrule(r){1-2} \cmidrule(lr){3-4} \cmidrule(lr){5-6} \cmidrule(l){7-8}
\textbf{Word}       & \textbf{Freq.}    & \textbf{Word}      & \textbf{Freq.}     & \textbf{Word}      & \textbf{Freq.}     & \textbf{Word}      & \textbf{Freq.}     \\ \midrule
judaslohn           & 58                & kemmerich          & 227                & corona             & 3019               & corona             & 2426               \\
qr                  & 36                & hanau              & 123                & coronavirus        & 611                & lockdown           & 301                \\
rheinruhrgebiet     & 35                & corona             & 123                & covid              & 380                & covid              & 276                \\
nameland            & 28                & qr                 & 101                & virolog            & 316                & coronavirus        & 162                \\
nazisau             & 24                & coronavirus        & 99                 & drost              & 236                & drost              & 162                \\
kassenbon           & 23                & nameland           & 97                 & maskenpflicht      & 229                & virolog            & 155                \\
wernerpilz          & 22                & himmelslatern      & 96                 & schutzmask         & 195                & maskenpflicht      & 124                \\
tagesvat            & 21                & affenhaus          & 78                 & jorgemario         & 180                & neuinfektion       & 101                \\
stockhaus           & 21                & soleimani          & 65                 & hamsterkauf        & 149                & streeck            & 87                 \\
schonau             & 20                & franknstein        & 58                 & sar                & 146                & jorgemario         & 78                 \\
horowitz            & 19                & nazisau            & 51                 & lockdown           & 117                & infektionszahl     & 77                 \\
veith               & 18                & malbuch            & 48                 & neuinfektion       & 108                & coronakris         & 71                 \\
kubitschek          & 17                & wuhan              & 36                 & jac                & 107                & sar                & 69                 \\
nowabo              & 17                & wellsow            & 35                 & coronakris         & 105                & schopendus         & 60                 \\
hollek              & 17                & sahnesteif         & 34                 & infektionszahl     & 81                 & kaufprami          & 54                 \\
floristin           & 16                & probost            & 30                 & wuhan              & 76                 & alteomar           & 52                 \\
wachleut            & 16                & hubertusprob       & 29                 & beatmungsgerat     & 75                 & qr                 & 50                 \\
fgm                 & 16                & hausdoc            & 27                 & greetz             & 75                 & testung            & 49                 \\
fuhrungsduo         & 16                & engstfeld          & 27                 & risikogebiet       & 74                 & partysz            & 45                 \\
gurkenland          & 15                & connewitz          & 26                 & ffp                & 71                 & nameland           & 41                 \\ \bottomrule
\end{tabular}
\begin{tablenotes}
      \scriptsize
      \item The words shown are preprocessed as described in Sec. \ref{sec:preprocessing} including lemmetization and lowercasing. Some words may also represent usernames that were frequently mentioned in comment threads. 
      \end{tablenotes}
\end{threeparttable}
\end{table}

\subsection{Temporal degradation test}

% \begin{figure}[htp]
%     \centering
%     \includegraphics[width=1.3\textwidth]{figures/double.001.jpeg}
%     \caption{Spearman correlation between months}
%     \label{fig:bar}
% \end{figure}

\begin{figure}
\centering
\begin{subfigure}{0.45\textwidth}
    \centering
    \includegraphics[width=1.0\textwidth]{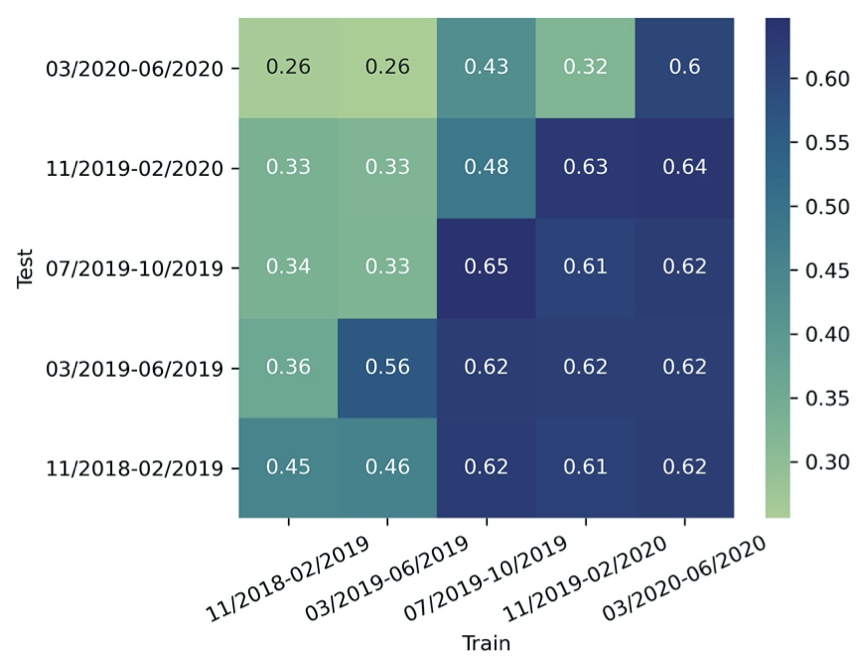}
    \caption{Comparison of F1-scores across four-month chunks}
    \label{fig:chunks}
\end{subfigure}
\begin{subfigure}{0.45\textwidth}
    \centering
    \includegraphics[width=1.0\textwidth]{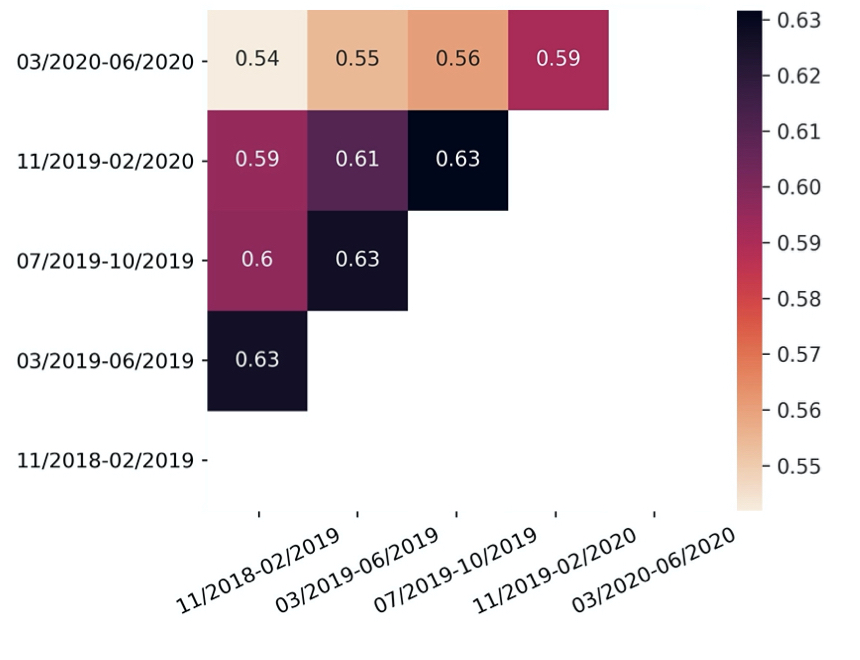}
    \caption{Spearman correlation across four-month chunks}
    \label{fig:spearman}
\end{subfigure}
\end{figure}

The results from the temporal degradation test show that, in general, as the time interval between the training and evaluation chunks increases, classifier performance decreases (Fig. \ref{fig:chunks}). Furthermore, the Spearman correlation between chunks (Fig. \ref{fig:spearman}) shows the same trend: as the time interval between two chunks increases, the corpora similarity decreases. 

However, the temporal degradation effect seen in Fig. \ref{fig:chunks} is significantly more potent in the forward direction of time where the training chunk precedes the evaluation chunk. We observe drops in the F1 score of up to 0.31 in the forward direction. In the backward direction of time where the time period of the training chunk is after the evaluation chunk, the temporal degradation effect is small or non-existent. One reason that the temporal degradation effect may be more negligible in the backward direction of time is that the news cycle has a "memory" where older stories are built upon, and the respective vocabulary is accumulated instead of being discarded.

The chunk from March 2020 to June 2020 is notable for both its low F1-scores in Fig. \ref{fig:chunks} and low Spearman correlation coefficients in Fig. \ref{fig:spearman}. The time period of this chunk coincides with the emergence of hundreds of COVID-19 related news stories and emerging new words like those in Table \ref{tab:words}. These results suggest that abrupt concept drift associated with the COVID-19 pandemic contributed to significant temporal degradation for classifiers trained on pre-pandemic data. 

\section{Discussion}
\label{sec:discussion}

The prevalence of abusive language in online comment sections presents a significant challenge for newspapers. As a result, there has been growing interest in (semi-) automated comment moderation tools that use machine learning and natural language processing to avoid the high costs of manual moderation or shutting down comments sections entirely. Unfortunately, however, much research on abusive language detection is modeled on an unrealistically static language environment where the language and abuse topics remain unchanging beyond the training dataset. 

This paper uses a time stratified evaluation procedure to show that the typical random train-test splitting strategy tends to overestimate classifier performance on future data. Random train-test splits assume a complete overlap between the training and testing data---an assumption that is broken as soon as the model is deployed into a real-world environment to make predictions on new data. We argue that a time-stratified evaluation procedure in which the training and testing data are selected from distinct time periods is better suited for modeling our real-world environment where language changes dynamically. 

Our findings on temporal degradation suggest that a classifier's performance can degrade significantly in as short a period as four months. Temporal degradation is especially pronounced during periods of abrupt concept drift like the COVID-19 pandemic. Our experiments consistently showed that the changes in vocabulary associated with the pandemic led to a sharp decline in performance among classifiers trained on data from before the pandemic. What is clear from these results, and other concept drift literature, is that the niave application of standard ML techniques will result in worse performance as the model's training corpus becomes increasingly outdated. 

Practitioners of ALD systems have several avenues available to deal with the consequences of temporal degradation and concept drifting data. On the low-tech end of the spectrum, are sliding-window training schemes in which models are regularly re-trained with new data and old data is discarded (see for example Nobata et al. 2016 \cite{nobata_abusive_2016}). Other, more complex adaption strategies are covered in Gama et al. 2014 \cite{gama_survey_2014}, though many of these algorithms are not yet available in standard ML libraries. In almost all cases, save for \textit{unsupervised} concept drift adaptation algorithms \cite{gemaque_overview_2020}, labeling some fraction of incoming data is required. This fact implies that manual moderation will continue to be necessary, even if the workload is greatly reduced. One promising direction to reduce the amount of new data that needs to be manually labeled is active learning \cite{settles.tr09}. Active learning queries data instances that are particularly useful for training (i.e., close to the decision boundary). 

Online discourse will likely be a permanent and growing fixture of how society communicates. Regulating online speech raises complex but important legal, political, and technical issues that have broad implications for interacting with media and forming opinions. If newspapers opt to use (semi-)automated comment moderation systems, it is crucial that these systems perform well and are aligned with the criteria for comment censorship outlined by the platform. Our findings support the notion that abusive language detection is not a trivial task. Classifiers trained to detect abusive language will need to be regularly updated with new data or otherwise designed to adapt to changes in language and the incoming data. Failure to do so risks ineffective comment moderation systems at best and careless censorship at worst.

%\section{Acknowledgments}
% Erst in finaler Version einfügen (sonst nicht blinded)	
%The research leading to these results received funding from the federal state of North Rhine-Westphalia and the European Regional Development Fund \linebreak (EFRE.NRW~2014-2020), Project: \raisebox{-.25\height}{\includegraphics[height=1.5\fontcharht\font`\B]{figures/moderat_logo_small}} (No.~CM-2-2-036a).

%We thank Dennis Assenmacher for his help with the machine learning work involved in this paper. 

% ----------------
% | Bibliography |
% ----------------

\bibliographystyle{splncs04}
\bibliography{./main.bib}
% ---------------------
% |  End of Document  |
% ---------------------

\end{document}